\documentclass[letterpaper]{article} 
\usepackage{aaai24}  
\usepackage{times}  
\usepackage{helvet}  
\usepackage{courier}  
\usepackage[hyphens]{url}  
\usepackage{graphicx} 
\usepackage{natbib}  
\usepackage{caption} 
\usepackage{textalpha}
\usepackage{algorithm}
\usepackage{algorithmic}

\usepackage{amsmath}
\usepackage{multirow}
\usepackage{makecell}
\usepackage{bbding}
\usepackage{graphicx}
\usepackage{adjustbox}
\usepackage{booktabs}
\usepackage{blindtext}
\usepackage{pifont}
\usepackage{newfloat}
\usepackage{listings}
\DeclareCaptionStyle{ruled}{labelfont=normalfont,labelsep=colon,strut=off} 
\floatstyle{ruled}
\newfloat{listing}{tb}{lst}{}
\floatname{listing}{Listing}
\title{Aleth-NeRF: Illumination Adaptive NeRF with Concealing Field Assumption}
\author{
    Ziteng Cui\textsuperscript{\rm 1,2},
    Lin Gu\textsuperscript{\rm 3,1},
    Xiao Sun\textsuperscript{\rm 2}\thanks{Corresponding Author},
    Xianzheng Ma\textsuperscript{\rm 4},
    Yu Qiao\textsuperscript{\rm 2}, 
    Tatsuya Harada\textsuperscript{\rm 1,3}
}
\affiliations{
    \textsuperscript{\rm 1}The University of Tokyo \\
    \textsuperscript{\rm 2}Shanghai AI Laboratory \\
    \textsuperscript{\rm 3}RIKEN AIP\\
    \textsuperscript{\rm 4}University of Oxford\\
    (cui, harada)@mi.t.u-tokyo.ac.jp, lin.gu@riken.jp, (sunxiao, maxianzheng)@pjlab.org.cn, yu.qiao@siat.ac.cn

}

\usepackage{bibentry}

\begin{document}

\maketitle

\begin{abstract}
The standard Neural Radiance Fields (NeRF) paradigm employs a viewer-centered methodology, entangling the aspects of illumination and material reflectance into emission solely from 3D points. This simplified rendering approach presents challenges in accurately modeling images captured under adverse lighting conditions, such as low light or over-exposure.
Motivated by the ancient Greek emission theory that posits visual perception as a result of rays emanating from the eyes, we slightly refine the conventional NeRF framework to train NeRF under challenging light conditions and generate normal-light condition novel views unsupervisedly. We introduce the concept of a ``Concealing Field,"
which assigns transmittance values to the surrounding air to account for illumination effects. In dark scenarios, we assume that object emissions maintain a standard lighting level but are attenuated as they traverse the air during the rendering process. 
Concealing Field thus compel NeRF to learn reasonable density and colour estimations for objects even in dimly lit situations. Similarly, the Concealing Field can mitigate over-exposed emissions during rendering stage. Furthermore, we present a comprehensive multi-view dataset captured under challenging illumination conditions for evaluation. Our code and proposed dataset are available at https://github.com/cuiziteng/Aleth-NeRF.
\end{abstract}

\begin{figure*}
    \centering
    \includegraphics[width=0.88\linewidth, height=10.3cm]{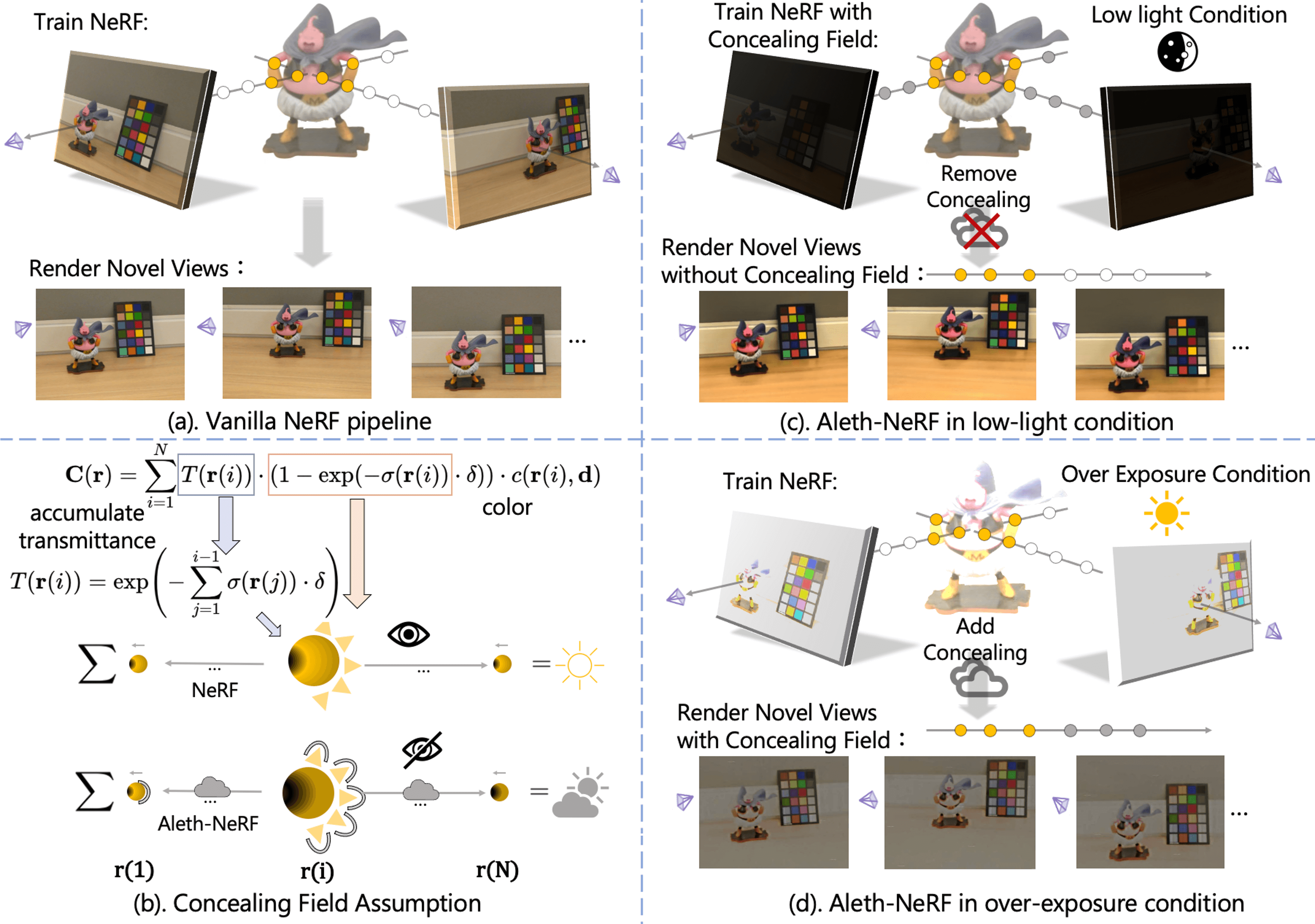}
    \caption{Utilizing the Concealing Field assumption, Aleth-NeRF is capable of processing both low-light $\&$ over-expose multi-view images as inputs and generating novel views with natural illumination.}
    \label{Fig:majin_buu}
\end{figure*}

\section{Introduction}
\label{sec:intro}

Neural Radiance Field (NeRF)~\cite{nerf} has been demonstrated to effectively understand 3D scenes from 2D images and generate novel views.
However, the formulation of NeRF and its follow-up variants assume captured images are under normal light, often failing to work under low-light~\cite{raw_nerf} or over-exposure scenarios. This is because vanilla NeRF is \textit{viewer-centered} which  models the amount of light emission from a location to the viewer without disentangling illumination and material (Fig.~\ref{Fig:majin_buu}(a))~\cite{lyu2022nrtf}.  As a result, the NeRF algorithm interprets a dark scene as insufficient radiation of the 3D object particles, violating the estimation of the object's material and geometry. In practical applications, images are often taken under challenging lighting conditions. Therefore, this paper aims to slightly modify vanilla NeRF  for under $\&$ over-exposure scenes.  As shown in Fig.~\ref{Fig:majin_buu}(c, d), the proposed, Aleth-NeRF,  renders normal-light novel views despite the severe input images.

 The rendering process in NeRF (Fig.~\ref{Fig:majin_buu}(b)) is similar to the \textit{viewer-centered}  emission theory held by ancient Greek. Emission theory ignores the incident light but postulates visual rays emitted from the eye travel in straight lines and interacts with objects to form the visual perception. Therefore, the darkness of an entity is solely caused by the particles between the object and the eye. In other words, all objects are visible by default unless concealed. Inspired by this worldview, we assume a simple but NeRF-friendly concept that it is the concealing fields   (gray particle in Fig.~\ref{Fig:majin_buu}(c)) in viewing direction that attenuates the emission and makes the viewer see a low-light scene. This is in contrast to the standard NeRF setting where the density of air   (white particle in Fig.~\ref{Fig:majin_buu}(a)) is usually zero. Introducing the Concealing Field, which assigns the air particles with transmittance value allows NeRF to accurately estimate the colour and density of objects (yellow particles in Fig.~\ref{Fig:majin_buu}(c)) in low-light conditions, therefore when removing the concealing fields, or Aletheia (\textalpha \textlambda\texteta\texttheta\textepsilon\textiota\textalpha)~\footnote{normally translated as "unconcealedness", "disclosure" or "revealing"~\cite{heidegger2010being}}, we are able to render novel views with normal-light. On the contrary, for the over-exposed scene, deliberately adding the concealing fields in rendering stage could correct the exposure.

Our proposed method Aleth-NeRF takes low-light $\&$ over-exposure images as inputs to train the model and learn the volumetric representation jointly with concealing fields. As shown in Fig.\ref{Fig:majin_buu}(b), we jointly train NeRF with concealing fields between the object and viewer. For  the low light scenario,  we  remove the concealing fields during the rendering stage  (Fig.~\ref{Fig:majin_buu}(c)). When dealing with over-exposure images, Aleth-NeRF would add concealing fields to suppress overly bright (Fig.~\ref{Fig:majin_buu}(d)). 
Our contributions are summarized as follow: 

\begin{itemize}
\item We propose \textbf{Aleth-NeRF}, that trains under low-light $\&$ over-exposure conditions and generates novel views under normal-light. Inspired by ancient Greek philosophy, we naturally extend the transmittance function in vanilla NeRF by modelling concealing fields between the objects and viewer to interpret lightness degradation.

    \item We contribute a challenging illumination multi-view dataset, with paired sRGB low-light $\&$ normal-light $\&$ over-exposure images, dataset would also be public.
    
    \item We compare  with various image enhancement and exposure correction methods  $\&$ previous NeRF-based method~\cite{raw_nerf}. Extensive experiments show that our \textbf{Aleth-NeRF} achieves satisfactory enhancement quality and multi-view consistency. 

\end{itemize}

\section{Related Works}
\label{sec:related_work}

\subsection{Novel View synthesis with NeRF}

NeRF~\cite{nerf} is proposed for novel view synthesis from a collection of posed input images. The unique advantage of NeRF models exists in preserving the 3D geometry consistency thanks to its physical volume rendering scheme. In addition several methods have been proposed to speed up and improve NeRF training~\cite{barron2021mipnerf,yu_and_fridovichkeil2021plenoxels,autoint2021,yu2021plenoctrees,NeRF_on_DieT,depth_nerf,mueller2022instant}. 

Many of the latter works focus on improving NeRF's performance under various degradation conditions, such as blurry~\cite{deblur_nerf}, noisy~\cite{Nan_2022_CVPR}, reflection~\cite{NERF_reflection}, glossy surfaces~\cite{refnerf_CVPR22}, underwater~\cite{seathru_nerf}, or use NeRF to handle super-resolution~\cite{wang2021nerf-sr,volume_sr} and HDR reconstruction~\cite{hdrnerf,jun2022hdr} in 3D space.
Another line of research extends NeRF for lightness editing in 3D space. Some work, like NeRF-W~\cite{nerf_wild}, focuses on rendering NeRF with uncontrolled in-the-wild images, other relighting works~\cite{srinivasan2021nerv,rudnev2022nerfosr,NeRFactor} rely on known illumination conditions and introduce additional physical elements (\textit{i.e.} normal, light, albedo, etc.), along with complex parametric modeling of these elements.
Meanwhile, these methods are not specifically designed for low-light $\&$ over-exposure conditions.

Among these, RAW-NeRF~\cite{raw_nerf} is more closer to our work, which proposes to render NeRF in HDR RAW domain and then post-process the rendered scene with image signal processor (ISP), RAW-NeRF has shown a preliminary ability to enhance the scene light but requires HDR RAW data for training, which make it hard to generalize on common used sRGB images. Instead our Aleth-NeRF could directly rendered on sRGB under $\&$ over exposure images and injection unsupervised enhancement into 3D space by an effective concealing fields manner.

\subsection{Enhancement in challenging light conditions}

 Challenging lightness can arise from multiple sources, encompassing natural lighting variances (such as low-light situations and overly bright scenes) as well as human-induced factors (such as incorrect camera exposure settings). To tackle these challenge lighting conditions, numerous techniques for image enhancement and exposure correction have been developed and proposed.

\subsubsection{Image Enhancement $\&$ Exposure Correction:} 
Image enhancement methods aims to enhance images with poor illumination, traditional methods usually rely on RetiNex theory~\cite{retinex,LIME} or Histogram Equalization~\cite{DIP_citation}, currently deep neural networks (DNNs) based methods become the mainstream solutions, series of CNN $\&$ Transformer-based methods have been developed~\cite{RetiNexNet,Deep_LPF,LLFlow,LLFormer,Enlightengan,Zero-DCE,ECCV22_jin2022unsupervised,SCI_CVPR2022,Yang_2023_ICCV}. Meanwhile, several exposure correction methods have been proposed to consider both under $\&$ over exposure conditions~\cite{Afifi_2021_CVPR,nsampi2021learning,BMVC22_IAT,Huang_2023_CVPR_exposure}, which aims to correct underexposure and its adverse overexposure images into normal-light condition.
However, image enhancement $\&$ exposure correction methods almost build on 2D image space operations, which often fail to exploit the 3D geometry of the scene and could not deal with multi-view inputs.

\subsubsection{Video Enhancement $\&$ Burst Enhancement:} 
Beyond above techniques that focus on single image. Video enhancement methods have been proposed to optimize the temporal consistency between adjacent frames, ensuring stability when processing different frames. These methods employ various approaches such as 3D convolution~\cite{Lv2018MBLLEN}, optical flow~\cite{LLVE_2021_CVPR}, and event guidance~\cite{video_enhance_event}.
Burst enhancement also plays a crucial role in modern computational photography area~\cite{mobile_photography,hasinoff2016burst}, where multiple frames are captured during exposure and processed using an ''align-merge-enhance" approach to produce a single output frame. In recent advancements, deep neural networks have been employed to replace traditional manual operation algorithms in these methods~\cite{burst_2018_ECCV,burst_KPN,burst_deep}.

However, existing image $\&$ video $\&$ burst enhancement methods primarily focus on enhancing images in their original views, rather than generating coherent 3D scenes with novel views, for comparison we have to combine these enhancement methods with NeRF (see Table.~\ref{table:LOM_results} and Table.~\ref{table:compare_exposure}). In contrast, Aleth-NeRF is capable of directly synthesizing novel views under challenging light conditions while achieving state-of-the-art enhancement quality.

\section{Methods}
\begin{figure}
    \centering
    \includegraphics[width=1.00\linewidth]{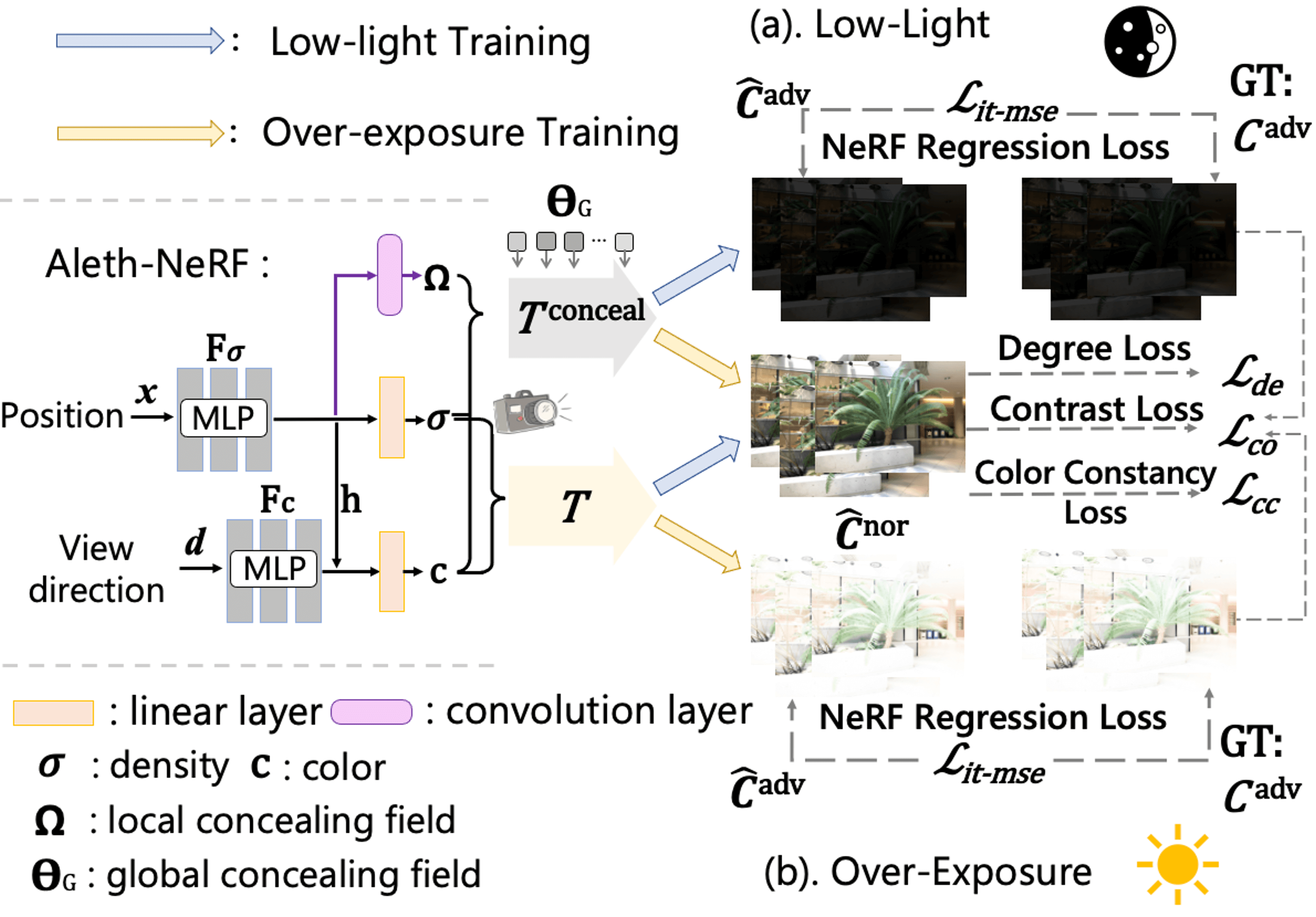}
    \caption{Train on adverse lighting condition images $C^{adv}$, Aleth-NeRF  performs unsupervised lightness correction by (a). remove concealing fields in low-light conditions and (b). add concealing fields in over-exposure conditions.}
    \label{fig:structure}
\end{figure}

\subsection{Neural Radiance Field Revisited}

Radiance Field is defined as the density $\sigma$ and RGB colour value $c$ of a 3D location $\textbf{x}$ under a 2D viewing direction $\textbf{d}$. The density $\sigma$, on the one hand, represents the radiation capacity of the particle itself at $\textbf{x}$, and on the other hand, controls how much radiance is absorbed when other lights pass through $\textbf{x}$.

When rendering an image, a camera ray $\textbf{r}(t) = \textbf{o} + t \cdot \textbf{d}$ ($\textbf{r} \in \textbf{R}$) cast from the given camera position $\textbf{o}$ towards direction $\textbf{d}$. All the radiance is accumulated along the ray to render its corresponding pixel value $\textbf{C}(\textbf{r})$. 
Formally, 
\begin{equation}
{\bf C}({\bf r}) = \int_{t_n}^{t_f} T(\textbf{r}(t))\sigma({\bf r}(t))c({\bf r}(t), {\bf d})dt,
\label{eq:volume_con}
\end{equation}
 where
\begin{equation}
T(\textbf{r}(t)) = \exp (- \int_{t_n}^{t} \sigma({\bf r}(s))ds),  
\label{eq:trans_con}
\end{equation}
is known as the \emph{accumulated transmittance} that denotes the radiance decay rate of the particle at $\textbf{r}(t)$ when it is occluded by particles closer to the camera (at $\textbf{r}(s)$, $s < t$).
The integrals are computed by a discrete approximation over sampled 3D points along the ray $\bf r$ (see Fig.~\ref{Fig:majin_buu}(b)). The discrete form of Eq.~\ref{eq:volume_con} and Eq.~\ref{eq:trans_con} are represented as follows:
\begin{equation}
\label{eq.volume_rendering}
{\bf C}({\bf r}) = \sum_{i=1}^N T(\textbf{r}(i))(1 - \exp(-\sigma(\textbf{r}(i)) \cdot \delta)) \cdot c(\textbf{r}(i), \textbf{d}),
\end{equation}
\begin{equation}
\label{eq.transmittance}
T(\textbf{r}(i)) = \exp \left(-\sum_{j=1}^{i-1}\sigma(\textbf{r}(j)) \cdot \delta \right). 
\end{equation}
 Value of  $\sigma(\textbf{r}(i))$ reflects the object occupancy in $\textbf{r}(i)$. $\delta$ is a constant distance value between adjacent sample points under uniform sampling.
\begin{figure}
    \centering
    \includegraphics[width=0.95\linewidth]{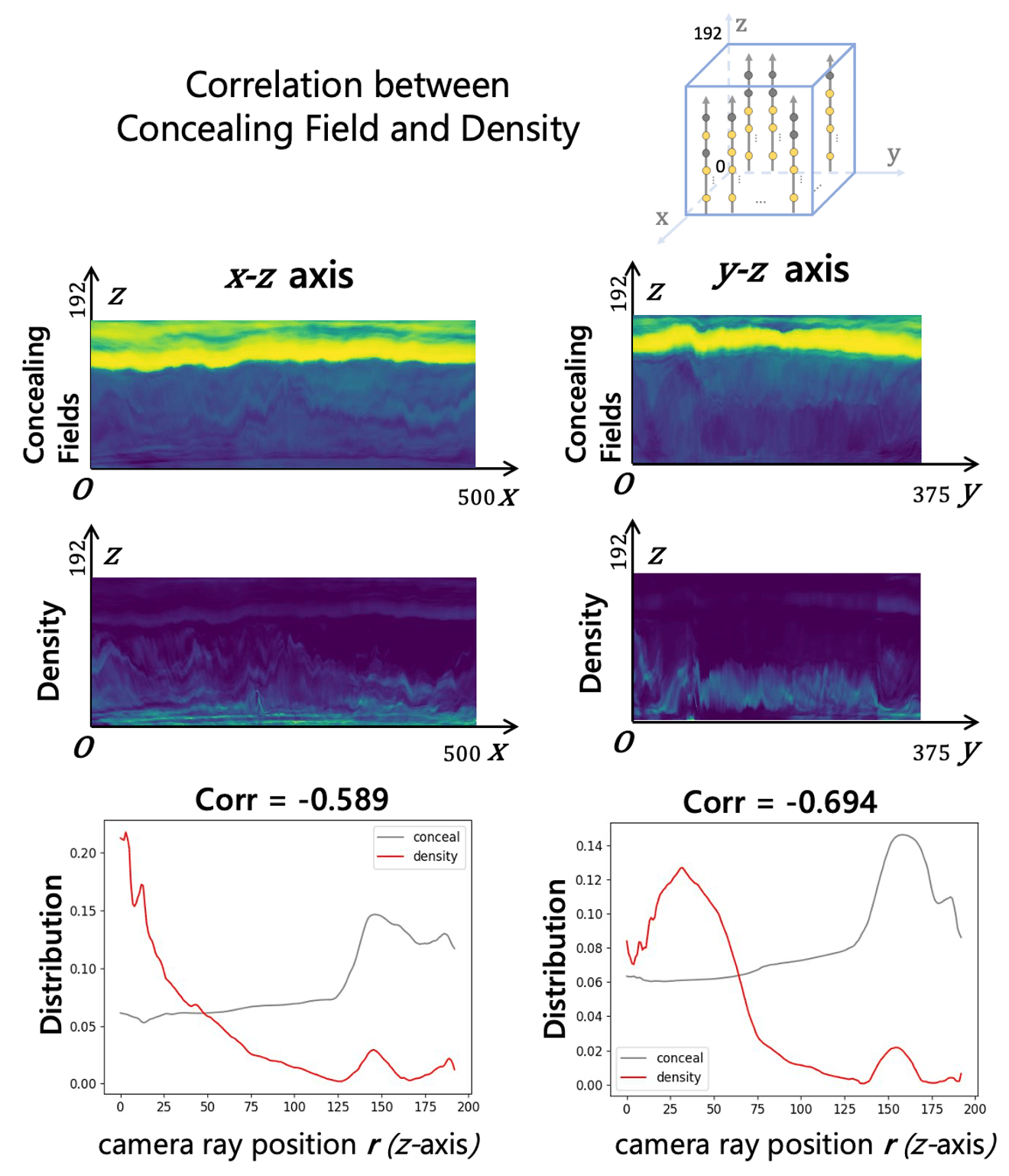}
    \caption{Along camera ray r ($z$ axis), concealing fields and density $\sigma$ exhibit a negative correlation, $(x, y)$ denotes training images' width and height.}
    \label{fig:corr}
\end{figure}

\begin{figure*}
    \centering
    \includegraphics[width=1.00\linewidth]{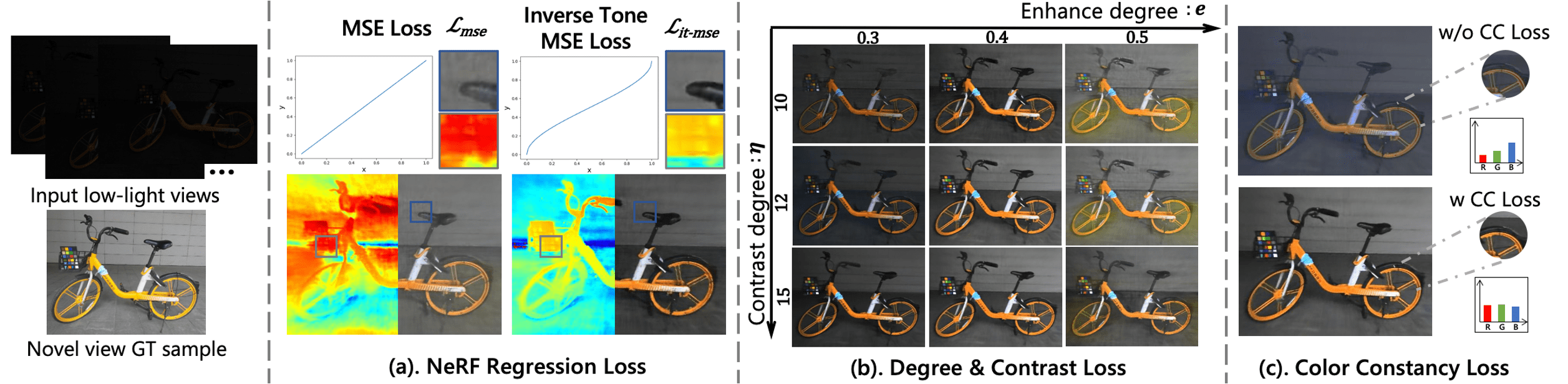}
    \caption{A low-light ``bike" scene for example, we show the ablation analyze of different loss functions' effectiveness.}
    \label{fig:loss_ablation}
\end{figure*}


For the network structure, NeRF learns two multilayer perceptron (MLP) networks: density MLP $F_{\sigma}$ and colour MLP $F_c$, to map the 3D location $\textbf{r}(i)$ and 2D viewing direction $\textbf{d}$ to its density $\sigma$ and colour $c$, specifically:
\begin{equation}
F_{\sigma}(\textbf{r}(i)) \rightarrow \sigma(\textbf{r}(i)), \textbf{h}
\end{equation}
\begin{equation}
F_c(\textbf{h}, \textbf{d}) \rightarrow c(\textbf{r}(i), \textbf{d}),
\end{equation}
where \textbf{h} is a hidden feature vector that is sent to colour MLP $F_c$, color $c(\textbf{r}(i), \textbf{d})$ and density $\sigma(\textbf{r}(i))$ are further activated by Sigmoid and ReLU functions to regularise their ranges into $[0, 1)$ and $[0, \infty)$ respectively.
Given ground truth images $\textbf{C}$, NeRF is optimised by minimising the MSE loss between predicted images $\hat{\textbf{C}}$ and ground truth images $\textbf{C}$:

\begin{equation}
    \mathcal L_{mse} = \sum_{\textbf{r}}^{\textbf{R}} || \hat{\textbf{C}}(\textbf{r}) - \textbf{C}(\textbf{r}) ||^2.
\label{eq:mse_loss}
\end{equation}
We refer more details such as positional encoding and hierarchical volume sampling to  NeRF paper~\cite{nerf}.

\subsection{Aleth-NeRF with Concealing Field}

Given adverse lighting condition images $\textbf{C}^{adv}(\textbf{r})$ taken in low-light $\&$ over-exposure conditions, our goal is to generate novel views $\textbf{C}^{nor}(\textbf{r})$ in normal light condition.
The key idea is that Aleth-NeRF assumes $\textbf{C}^{adv}$ and $\textbf{C}^{nor}$ are rendered with the same underlying density field $\sigma$ (yellow particle in Fig.~\ref{Fig:majin_buu}) along each camera ray $\textbf{r}$, but with or without the proposed concealing fields (gray particle in Fig.~\ref{Fig:majin_buu}).

We model two types of concealing fields to reduce light transport in the volume rendering stage: local Concealing Field denoted as $\Omega$ at the voxel level, and global Concealing Field denoted as $\Theta_G$ at the scene level. Multiplying  two fields, $\Omega$ and $\Theta_G$  gives the final concealing field value which reduce the \textit{accumulated transmittance}. Fig.~\ref{fig:structure} shows an overview of the Aleth-NeRF training strategy. Fig.~\ref{fig:corr} shows concealing fields $\Omega \cdot \Theta_G$'s distribution.

\subsubsection{Local Concealing Field} denoted as $\Omega(\textbf{r}(i))$, defines an extra light concealing capacity of a particle at 3D location $\textbf{r}(i)$. As depicted in Fig.~\ref{fig:structure}, $\Omega$ is individually learned for each 3D position and is generated using the density MLP $F_{\sigma}$. To create the local concealing field $\Omega$, we introduce an additional large kernel convolution layer (with a size of 7) built upon $F_{\sigma}$. This convolution process establishes spatial relationships between pixels and enriches the Concealing Field with predominantly light-related information rather than structural information~\cite{Zamir2020CycleISP}. This larger kernel convolution also effectively suppresses noise and contributes to smoother rendering outcomes.

\begin{equation}
conv_{size:7}(F_{\sigma}(\textbf{r}(i))) \rightarrow \Omega(\textbf{r}(i))
\end{equation}

\subsubsection{Global Concealing Field} denoted as $\Theta_G(i)$, is defined as a set of learnable parameters corresponding to the camera distance $i$ for all camera rays in $\textbf{R}$. $\Theta_G$ remains constant within a scene and is independent of voxels, as we posit that a specific degree of lighting influence remains consistent across the same scene. In our experiments, we initialize $\Theta_G(i)$ with a value of 0.5 for each camera distance $i$. With the global and local concealing fields, the \emph{accumulated transmittance} $T$ in Eq.~\ref{eq.transmittance} is blighted by $\Omega$ and $\Theta_G$ to mimic the process of light suppression (see Fig.~\ref{Fig:majin_buu}(b)), as follow:

\begin{equation}
\begin{aligned}
\label{eq.transmittance_local_global}
T^{conceal}(\textbf{r}(i)) = & \exp \left(-\sum_{j=1}^{i-1}\sigma(\textbf{r}(j)) \cdot  \delta \right) \\ 
& \cdot \prod_{j=1}^{i-1} \Omega(\textbf{r}(j)) \Theta_G(j).
\end{aligned}
\end{equation}

\subsubsection{Train and Test Schemes} of Aleth-NeRF for both low-light and over-exposure scenes are illustrated in Fig.~\ref{fig:structure}. For low-light conditions (blue arrow in Fig.~\ref{fig:structure}), Aleth-NeRF incorporates $T^{conceal}$ during the volume rendering stage to generate low-light images $\hat{\textbf{C}}^{adv}$, while removing the concealing fields to produce normal-light images $\hat{\textbf{C}}^{nor}$ with the original $T$ in Eq.~\ref{eq.transmittance}. 
Conversely, for over-exposure scenes (yellow arrow in Fig.~\ref{fig:structure}), Aleth-NeRF employs $T$ to render very-bright images $\hat{\textbf{C}}^{adv}$ and incorporates the concealing fields along with $T^{conceal}$ to render normal-light images $\hat{\textbf{C}}^{nor}$. The NeRF regression loss is applied to compare predicted adverse-light images $\hat{\textbf{C}}^{adv}$ with the corresponding ground truth $\textbf{C}^{adv}$, while unsupervised lightness correction losses ensure the generation of predicted normal-light images $\hat{\textbf{C}}^{nor}$, more details please refer to next section.

\subsubsection{Analysis of Concealing Fields} is shown in Fig.~\ref{fig:corr}, we analyse the distribution of concealing fields and density along the camera ray $\textbf{r}(i)$ ($z$ axis), where $x$, $y$ are the training images' width and height. In Fig.~\ref{fig:corr}, brighter region denotes the larger possibility there exists concealing fields (gray line) or density (red line). We find that concealing fields exist in location $\textbf{r}(i)$ with lower density value $\sigma(\textbf{r}(i))$. And the Pearson correlation coefficient $\textbf{Corr}$ between concealing fields and density $\sigma$ are also negative along both $x-z$ axis (-0.589) and $y-z$ axis (-0.694). This validates that concealing fields are separated from density, thus rarely participating in scene rendering. Concealing fields exists more in locations $\textbf{r}(i)$ with sparse density, \textit{i.e.} air outside the objects.

\subsection{Effective Losses for Unsupervised Training}
\label{sec.priors}
In this section, we present the loss functions that guide the unsupervised lightness correction of Aleth-NeRF, facilitating improve novel views synthesis and enhancement in low-light $\&$ over-exposure conditions. An overview of the unsupervised training strategy is shown in Fig.~\ref{fig:structure}.

\subsubsection{NeRF Regression Loss:}
Direct apply NeRF's MSE loss function $\mathcal L_{mse}$ in low-light and over-bright scenes would result in an imbalance of pixel weights. Dark pixels with small values would contribute minimally, while bright pixels with larger values would dominate the contribution. To rectify this imbalance and ensure effective NeRF training under non-standard lighting conditions, we propose the incorporation of an inverse tone curve~\cite{brooks2019unprocessing,Cui_2021_ICCV} denoted as $\Phi$. This curve serves to re-balance pixel weights, additionally we introduce a small value $\epsilon = 1e^{-3}$ within the regression loss to facilitate more accurate estimations. As a result, the formulated inverse tone MSE  loss  $\mathcal L_{it-mse}$ for NeRF training is presented as follows:
 
\begin{equation}
\begin{aligned}
    &\Phi(x) = \frac{1}{2} - sin(\frac{sin^{-1}(1-2x)}{3}), \\
    & \mathcal L_{it-mse} = \sum_{\textbf{r}}^{\textbf{R}} || \Phi(\hat{\textbf{C}}(\textbf{r}) + \epsilon) - \Phi(\textbf{C}(\textbf{r}) + \epsilon) ||^2,
\label{eq:it-mse_loss}
\end{aligned}
\end{equation}
the comparison of $\mathcal L_{mse}$ and $\mathcal L_{it-mse}$ is shown in Fig.~\ref{fig:loss_ablation}(a), inverse tone MSE loss $\mathcal L_{it-mse}$ can enhance scene clarity while also 
suppress noise to a certain extent.

\subsubsection{Degree $\&$ Contrast Loss:}
In order to regulate the level of enhancement, we incorporate a degree loss denoted as $\mathcal L_{de}$ on the predict normal-light images $\hat{\textbf{C}}^{nor}$. $\mathcal L_{de}$ involves a hyperparameter $\mathbf{e}$, which signifies the enhancement degree inherent to Aleth-NeRF. To compute $\mathcal L_{de}$, a global average pooling operation is applied to the $\hat{\textbf{C}}^{nor}$ of each training batch. Subsequently, we minimize the degree loss $\mathcal L_{de}$ by comparing the pooled $\hat{\textbf{C}}^{nor}$ with the specified degree value $\mathbf{e}$. We set $\mathbf{e}$ to 0.45 in all experiments, ablation analyse of $\mathbf{e}$ is in Fig.~\ref{fig:loss_ablation}(b).

\begin{equation}
    \mathcal L_{de} = || \rm{avgpool}(\hat{\textbf{C}}^{nor}(\textbf{r})) - \mathbf{e}||^2.
\end{equation}

Then we additional design a contrast degree loss $\mathcal L_{co}$ between the adjacent sampling pixels of predict normal-light images $\hat{\textbf{C}}^{nor}$ and GT adverse-light images $\textbf{C}^{adv}$:

\begin{equation}
\label{eq.contrast_degree}
\begin{aligned}
     \mathcal L_{co} = & \sum_{k \in [+1, -1]} (\hat{\textbf{C}}^{nor} (\textbf{r}) - \hat{\textbf{C}}^{nor} (\textbf{r}+k)) - \\ & 
 \mathbf{e} \cdot \eta \cdot (\textbf{C}^{adv} (\textbf{r}) - \textbf{C}^{adv} (\textbf{r}+k)),
\end{aligned}
\end{equation}
the contrast degree loss would maintain the structure similarity between $\hat{\textbf{C}}^{nor}$ and $\textbf{C}^{adv}$, meanwhile control the contrast degree of $\hat{\textbf{C}}^{nor}$ by parameter $\eta$, ablation analyze of $\eta$ in low-light scene is shown in Fig.~\ref{fig:loss_ablation}, when enhance degree $\mathbf{e}$ is fixed, higher $\eta$ represents higher contrast in $\hat{\textbf{C}}^{nor}$. While in over-exposure scene, $\mathbf{e} \cdot \eta$ would fixed to $0.5$.

\subsubsection{Color Constancy Loss:}  
Photos taken in low-light $\&$ over exposure conditions would easily lose their color information, directly remove $\&$ add the concealing fields may cause color imbalance. To regularize the color of the predict normal-light images $\hat{\textbf{C}}^{nor}$, we introduce color constancy loss $\mathcal L_{cc}$ on $\hat{\textbf{C}}^{nor}$. Here we assume that $\hat{\textbf{C}}^{nor}$ obey the gray-world assumption~\cite{gray_world,Zero-DCE,ECCV22_jin2022unsupervised}, as follows:

\begin{equation}
    \mathcal L_{cc} = \sum_{p, q} (\hat{\textbf{C}}^{nor} (\textbf{r})^p - \hat{\textbf{C}}^{nor} (\textbf{r})^q)^2,
\end{equation}
where $(p, q) \in \{ (R, G), (G, B), (B, R) \}$ represents any pair of color channels, an ablation of color constancy loss $\mathcal L_{cc}$ can be found in Fig.~\ref{fig:loss_ablation}(c).

Above all, Aleth-NeRF's loss function $\mathcal L$ comprises four components: NeRF rendering loss $\mathcal L_{it-mse}$, along with three unsupervised lightness correction losses ($\mathcal L_{de}$, $\mathcal L_{co}$, and $\mathcal L_{cc}$). The overall training loss is then represented as:

\begin{equation}
    \mathcal L = \mathcal L_{it-mse} + \lambda_1 \cdot \mathcal L_{de} + \lambda_2 \cdot \mathcal L_{co} + \lambda_3 \cdot \mathcal L_{cc},
\end{equation}
where $\lambda_1$, $\lambda_2$ and $\lambda_3$ are three non-negative parameters to balance total loss weights, which we set to $1e^{-3}$, $1e^{-3}$ and $1e^{-8}$ respectively.

\begin{figure}[t]
    \centering
    \includegraphics[width=0.95\linewidth]{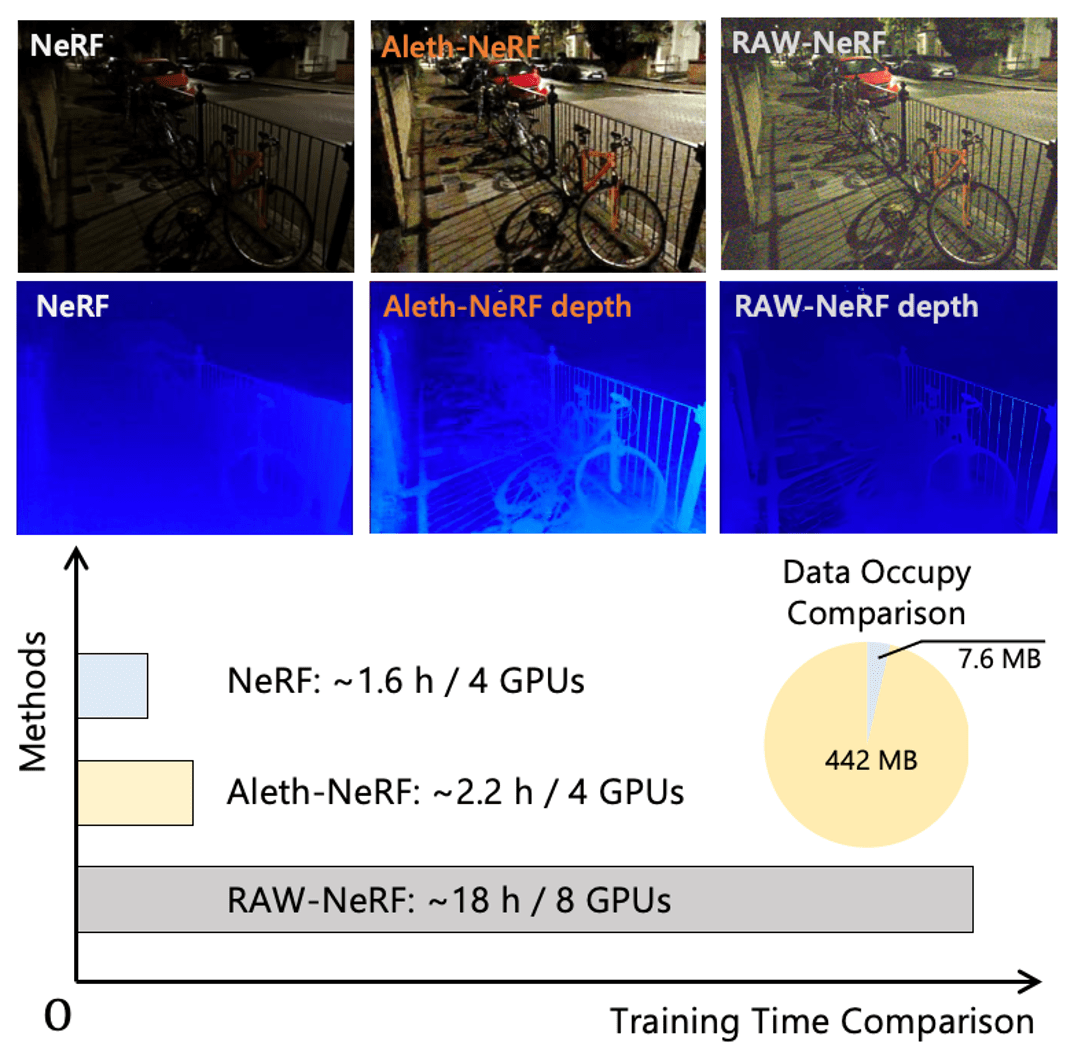}
    \caption{An comparison of enhancement results and model efficiency with RAW-NeRF~\cite{raw_nerf}, note that RAW-NeRF take 16-bits HDR image as inputs while NeRF $\&$ Aleth-NeRF take 8-bits LDR image as inputs.}
    \label{fig:RAW-NeRF}
\end{figure}

\section{Experiments}


In this section, we first introduce our collected dataset with various exposure paired multi-view images, then we present the novel view synthesis results of Aleth-NeRF under both low-light and under-exposure settings. For practical implementation, our framework builds on the open-source PyTorch toolbox \texttt{NeRF-Factory}~\footnote{https://github.com/kakaobrain/nerf-factory}. We utilize the Adam optimizer with an initial learning rate of $5e^{-4}$ and employ a cosine learning rate decay strategy every 2500 iterations. The training batch size is set at 4096 for a total of 62500 iterations. To ensure fairness, all comparative experiments within the proposed dataset use the same training strategies and parameter configurations.

\subsection{Challenging Illumination Multi-view Dataset}
\label{exp:dataset}
In this section, we introduce our collected paired low-light $\&$ normal-light $\&$ over-exposure multi-view dataset, names LOM dataset. Previous work~\cite{raw_nerf} also includes some low-light scenes, but their dataset concentrates on RAW denoising rather than sRGB low-light enhancement and does not include normal-light sRGB ground truth, making it hard to evaluate enhancement performance in novel view synthesis.

\begin{table}[t]
\renewcommand\arraystretch{1.33}
\centering
\begin{adjustbox}{max width = 1.00\linewidth}
\begin{tabular}{c|ccccc}
\hline
\hline
scene            & ``\textbf{\textit{buu}}" & ``\textbf{\textit{chair}}" & ``\textbf{\textit{sofa}}" & ``\textbf{\textit{bike}}" & ``\textbf{\textit{shrub}}" \\ \hline
\hline
collected views  & 25  & 48    & 33   & 40   & 35    \\ \hline
training views   & 22  & 43    & 29   & 36   & 30    \\ \hline
evaluation views & 3   & 5     & 4    & 4    & 5     \\ \hline
\hline
\end{tabular}
\end{adjustbox}
\caption{Details of the dataset split for LOM.}
\label{data_split}
\end{table}

In our proposed LOM dataset, we collected 5 scenes  (``\textbf{\textit{buu}}", ``\textbf{\textit{chair}}", ``\textbf{\textit{sofa}}", ``\textbf{\textit{bike}}", ``\textbf{\textit{shrub}}") in real-world. Each scene includes $25 \sim 48$ images, for each view's image, we generate low-light $\&$ normal-light $\&$ over-exposure pairs by adjusting exposure time and ISO while other configurations of the camera are fixed. We capture multi-view images by moving and rotating the camera tripod. Images are collected with resolution 3000 $\times$ 4000. We down-sample the original resolution with ratio 8 to 375 $\times$ 500 for convenience, and generate the ground truth view and angle information by adopting COLMAP~\cite{COLMAP1,COLMAP2} on the normal-light part images. For dataset split, in each scene, we choose 3 $\sim$ 5 images as the testing set, 1 image as the validation set, and other images to be the training set, details of training and evaluation views split is shown in Table.~\ref{data_split}. More details of our dataset please refer to supplementary part.

\begin{figure*}
    \centering
    \includegraphics[width=1.00\linewidth, height=11.5cm]{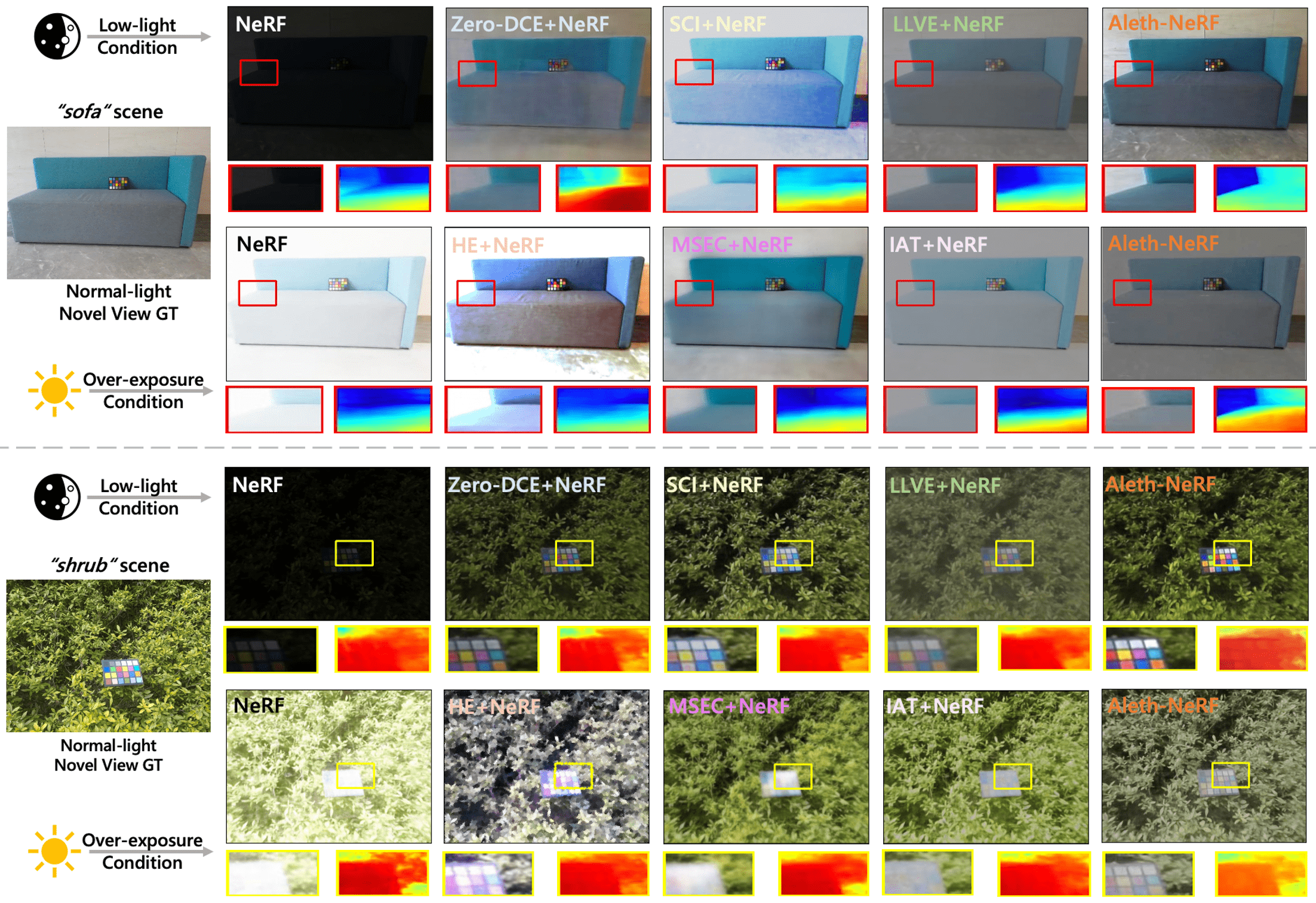}
    \caption{Novel view synthesis comparison in low-light and over-exposure conditions.}
    \label{fig:comparision}
\end{figure*}

\begin{table*}[] 
\large
\renewcommand\arraystretch{1.5}
\centering

\begin{adjustbox}{max width = 1.00\linewidth}
\fontsize{11}{9}
\begin{tabular}{l|c|c|c|c|c|c}

\toprule
\toprule
\multirow{2}{*}{\large{Method}} & ``\textbf{\textit{buu}}"              & ``\textbf{\textit{chair}}"              & ``\textbf{\textit{sofa}}"                & ``\textbf{\textit{bike}}"                & ``\textbf{\textit{shrub}}"               & \textbf{\textit{mean}}                \\ \cline{2-7} 
                        & PSNR/ SSIM/ LPIPS  & PSNR/ SSIM/ LPIPS   & PSNR/ SSIM/ LPIPS   & PSNR/ SSIM/ LPIPS   & PSNR/ SSIM/ LPIPS   & PSNR/ SSIM/ LPIPS   \\ \hline 
\large{NeRF}                    & \enspace 7.51/ 0.291/ 0.448  & \enspace 6.04/ 0.147/ 0.594  & 6.28/ 0.210/ 0.568  & \enspace 6.35/ 0.072/ 0.623  & \enspace 8.03/ 0.031/ 0.680  & \enspace 6.84/ 0.150/ 0.582  \\ \hline

\multicolumn{7}{c}{\ding{172} NeRF + Image Enhancement Methods} \\\hline

NeRF + RetiNexNet      & 11.64/ 0.646/ 0.395 & 11.04/ 0.631/ 0.561 & 12.85/ 0.738/ 0.527 & 17.79/ 0.653/ 0.550 & 11.85/ 0.211/ 0.583 & 13.03/ 0.576/ 0.523 \\
NeRF + Zero-DCE              & 17.81/ 0.833/ 0.357 & 12.44/ 0.684/ 0.547 & 14.43/ 0.787/ 0.539 & 10.16/ 0.468/ 0.557 & 12.58/ 0.282/ 0.540 & 13.48/ 0.610/ 0.488  \\
\large{NeRF + SCI}              & 7.84/ 0.660/ 0.562 & 12.07/ 0.699/ 0.584 & 10.25/ 0.737/ 0.626 & \underline{18.84}/ 0.637/ 0.565 & 12.38/ 0.358/ 0.587 & 12.27/ 0.618/ 0.585 \\
\large{NeRF + IAT}              & 14.03/ 0.656/ 0.453 & 19.08/ 0.800/ 0.565 & 10.49/ 0.528/ 0.678 & 17.16/ 0.657/ 0.534 & 16.15/ 0.344/ 0.573 & 15.38/ 0.597/ 0.561 \\ \hline 

\multicolumn{7}{c}{\ding{173} Image Enhancement Methods + NeRF} \\\hline

\large{RetiNexNet + NeRF}       & 16.19/ 0.780/ 0.396 & 16.89/ 0.756/ 0.543 & 16.98/ 0.807/ 0.577 & 18.00/ \underline{0.707}/ 0.482 &    14.86/ 0.284/ 0.518     &    16.58/ 0.667/ 0.503   \\
\large{Zero-DCE + NeRF}             & 17.90/ 0.858/ 0.376 & 12.58/ 0.721/ \textbf{0.460} & 14.45/ 0.831/ \underline{0.419} & 10.39/ 0.518/ \underline{0.464} & 12.32/ 0.308/ 0.481 & 13.53/ 0.649/ \underline{0.432} \\
\large{SCI + NeRF}              & 7.76/ 0.692/ 0.525 &    19.77/ \underline{0.802}/ 0.674  & 10.08/ 0.772/ 0.520 & 13.44/ 0.658/ \textbf{0.435}  &              \underline{18.16}/ \underline{0.503}/ \underline{0.475}        &   13.84/ 0.689/ 0.510 \\
\large{IAT + NeRF}              & 14.46/ 0.705/ 0.386 & 18.70/ 0.780/ 0.665 & 17.88/ 0.829/ 0.547 & 13.65/ 0.616/ 0.528 & 13.87/ 0.317/ 0.536 & 15.71/ 0.649/ 0.532 \\ 
\hline 

\multicolumn{7}{c}{\ding{174} Video Enhancement Methods + NeRF} \\\hline
\large{MBLLEN + NeRF}               & \textbf{22.39}/ \textbf{0.877}/ \underline{0.353} & \textbf{23.59}/ 0.788/ 0.559 & \textbf{19.99}/ 0.836/ 0.542 & 14.09/ 0.636/ 0.525 & 13.17/ 0.501/ 0.555 & \underline{18.65}/ \underline{0.728}/ 0.507 \\
\large{LLVE + NeRF}               & 19.97/ 0.848/ 0.393 & 15.17/ 0.764/ 0.610 & 18.17/ \underline{0.855}/ 0.465 & 13.84/ 0.638/ 0.492 & 15.35/ 0.287/ 0.577 & 16.50/ 0.678/ 0.507 \\
\hline \multicolumn{7}{c}{\ding{175} Our Proposed End-to-end Method} \\\hline
\large{\textbf{Aleth-NeRF}}              & \underline{20.22}/ \underline{0.859}/ \textbf{0.315} & \underline{20.93}/ \textbf{0.818}/ \underline{0.468} & \underline{19.52}/ \textbf{0.857}/ \textbf{0.354} & \textbf{20.46}/ \textbf{0.727}/ 0.499 & \textbf{18.24}/ \textbf{0.511}/ \textbf{0.448} & \textbf{19.87}/ \textbf{0.754}/ \textbf{0.417} \\ \bottomrule \bottomrule
\end{tabular}
\end{adjustbox}
\caption{We assess novel view synthesis performance in low-light settings by comparing generated images with ground truth normal-light views. Our evaluation metrics include PSNR $\uparrow$, SSIM $\uparrow$ and LPIPS $\downarrow$. Bold denotes the best result, Underline denotes the second best result.}
\label{table:LOM_results}
\end{table*}

\begin{table*}[] 
\large
\renewcommand\arraystretch{1.5}
\centering

\begin{adjustbox}{max width = 1.00\linewidth}
\fontsize{11}{9}
\begin{tabular}{l|c|c|c|c|c|c}

\toprule
\toprule
\multirow{2}{*}{\large{Method}} & ``\textbf{\textit{buu}}"              & ``\textbf{\textit{chair}}"              & ``\textbf{\textit{sofa}}"                & ``\textbf{\textit{bike}}"                & ``\textbf{\textit{shrub}}"               & \textbf{\textit{mean}}                \\ \cline{2-7} 
                        & PSNR/ SSIM/ LPIPS  & PSNR/ SSIM/ LPIPS   & PSNR/ SSIM/ LPIPS   & PSNR/ SSIM/ LPIPS   & PSNR/ SSIM/ LPIPS   & PSNR/ SSIM/ LPIPS   \\ \hline 
\large{NeRF}                    & \enspace 7.12/0.674/0.499  & 11.05/ 0.741/ \textbf{0.418}  & 10.22/ 0.783/ 0.475  & \enspace 9.65/ 0.698/ 0.416  & \enspace 9.96/ 0.405/ \underline{0.480}  & \enspace 9.60/ 0.660/ \underline{0.457}  \\ \hline

\multicolumn{7}{c}{\ding{172} NeRF + Exposure Correction Methods} \\\hline

NeRF + HE              & 14.34/ 0.613/ 0.673 & 15.37/ 0.661/ 0.590 & 16.69/ 0.733/ 0.558 & 15.32/ 0.627/ 0.458 & 11.97/ 0.468/ 0.556 & 14.74/ 0.620/ 0.567  \\
NeRF + IAT              & 14.11/ 0.780/ \underline{0.433} & \underline{19.24}/ \underline{0.810}/ 0.491 & 16.60/ 0.837/ \underline{0.459} & 17.73/ 0.760/ 0.394 & 14.05/ 0.381/ 0.499 & 16.35/ 0.714/ \textbf{0.455} \\
NeRF + MSEC              & 16.13/ 0.800/ \textbf{0.427} & 15.60/ 0.786/ \underline{0.472} & 16.56/ 0.807/ 0.495 & 12.60/ 0.716/ 0.465 & 13.66/ 0.332/ 0.509 & 14.91/ 0.688/ 0.473 \\ \hline 

\multicolumn{7}{c}{\ding{173} Exposure Correction Methods + NeRF} \\\hline

\large{HE + NeRF}       & 14.65/ 0.743/ 0.519 & 15.55/ 0.736/ 0.497 & 17.11/ 0.781/ 0.477 & 15.77/ 0.692/ \textbf{0.367} &    12.61/ \textbf{0.506}/ 0.622     &    15.13/ 0.692/ 0.496   \\
\large{IAT + NeRF}             & \underline{16.22}/ \underline{0.815}/ 0.486 & 18.98/ 0.799/ 0.503 & \underline{18.45}/ \underline{0.849}/ 0.478 & \underline{19.63}/ \textbf{0.776}/ 0.408 & \underline{15.63}/ 0.434/ \textbf{0.477} & \underline{17.78}/ \underline{0.734}/ 0.470 \\
\large{MSEC + NeRF}              & 15.53/ \textbf{0.817}/ 0.499 & 16.95/ 0.758/ 0.580 & \textbf{19.60}/ 0.817/ 0.498 & 18.90/ 0.725/ 0.483 & 15.48/ 0.400/ 0.499 & 17.29/ 0.703/ 0.512 \\ 
\hline 

\hline \multicolumn{7}{c}{\ding{174} Our Proposed End-to-end Method} \\\hline
\large{\textbf{Aleth-NeRF}}              & \textbf{16.78}/ 0.805/ 0.611 & \textbf{20.08}/ \textbf{0.820}/ 0.499 & 17.85/ \textbf{0.852}/ \textbf{0.458} & \textbf{19.85}/ \underline{0.773}/ \underline{0.392} & \textbf{15.91}/ \underline{0.477}/ 0.483 & \textbf{18.09}/ \textbf{0.745}/ 0.488 \\ \bottomrule \bottomrule
\end{tabular}
\end{adjustbox}
\caption{We assess novel view synthesis performance in over-exposure settings by comparing generated images with ground truth normal-light views. Our evaluation metrics include PSNR $\uparrow$, SSIM $\uparrow$ and LPIPS $\downarrow$. Bold denotes the best result, Underline denotes the second best result.}
\label{table:compare_exposure}
\end{table*}

\subsection{Novel View Synthesis in Low-light Condition}
\label{exp:multi_view}
In this section, we show novel view synthesis results in low-light scenes, and make comparison between predicted normal-light views with ground truth normal-light views (see Table.~\ref{table:LOM_results}). We design multiple comparison experiments in proposed dataset as follow:

We start by training vanilla NeRF~\cite{nerf} under low-light conditions as a baseline (first row in Table~\ref{table:LOM_results}). \ding{172} Subsequently, we apply various state-of-the-art enhancement methods to post-process the rendered NeRF results (denote as ``NeRF + " in Table~\ref{table:LOM_results}). \ding{173} Then we pre-process low-light images with enhancement methods and then train NeRF on these enhanced images (denote as ``* + NeRF" in Table.~\ref{table:LOM_results}). Notably, we observe that the ``NeRF + *" methods tend to underperform compared to the ``* + NeRF" methods in low-light setting, especially in perceptual similarity (LPIPS $\downarrow$). This potentially arise from NeRF's production of poor-quality images in dark scenes, causing enhancement methods to struggle when applied to the generated dark scene by NeRF. \ding{174} At last, we apply two video enhancement methods MBLLEN~\cite{Lv2018MBLLEN} and LLVE~\cite{LLVE_2021_CVPR} to pre-process low-light images and then trained with NeRF. \ding{175} Above all, our Aleth-NeRF could gain the best or second best performance on most scenes (see Table.~\ref{table:LOM_results}), at the same time Aleth-NeRF does not require any training data as prior knowledge, and trained in an end-to-end manner without any pre-processing nor post-processing.

Beyond  enhancement methods, RAW-NeRF~\cite{raw_nerf} also conducted experiments under low-light conditions, due to their model is trained on RAW data and hard to implement on our dataset, a comparison with RAW-NeRF on their proposed dataset scene is shown in Fig.~\ref{fig:RAW-NeRF}, while NeRF and Aleth-NeRF are trained with 8-bits LDR inputs and RAW-NeRF is trained with 16-bits HDR inputs, our Aleth-NeRF could also gain delight enhancement results while save much data occupy and training time.

Fig.~\ref{fig:comparision} shows qualitative visualization results in LOM dataset's ``\textbf{\textit{sofa}}'' and ``\textbf{\textit{shrub}}'' scenes, with the comparison of current SOTA image enhancement methods Zero-DCE~\cite{Zero-DCE}, SCI~\cite{SCI_CVPR2022} and video enhancement method LLVE~\cite{LLVE_2021_CVPR}, we can see that our Aleth-NeRF could gain more vivid enhancement results and closer to ground truth, meanwhile also gain much better in multi-view consistency (see depth maps). More visualization results please refer to supplementary part.

\subsection{Novel View Synthesis in Over-Exposure Condition}

Then we show novel view synthesis results in
over-exposure scenes, similar to low-light scenes, we make comparison with 3 mainstream exposure correction methods, including traditional histogram equalization (HE) method~\cite{DIP_citation} and two current SOTA deep network methods MSEC~\cite{Afifi_2021_CVPR} and IAT~\cite{BMVC22_IAT}. 

As it shown in Table.~\ref{table:compare_exposure}, we use vanilla NeRF trained on over-exposure images to serve as a baseline. \ding{172} Then we adopt exposure correction methods to post-process the generated results of NeRF (denote as ``NeRF + *"). \ding{173} We also adopt exposure correction methods to pre-process over-exposure images and then train NeRF on the processed images (denote as ``* + NeRF"). \ding{174} Our proposed Aleth-NeRF method. could gain best results in PSNR and SSIM, but the performance in LPIPS is not very good. 
Some visualization results are shown in Fig.~\ref{fig:comparision}, our Aleth-NeRF could suppress over-bright meanwhile maintain multi-view consistency.

One major reason for the less pronounced LPIPS results on over-exposed images  maybe that Aleth-NeRF’s training directly based on degraded overexposed images. Therefore, unsupervised learning faces challenges in compensating for these degraded pieces of information. Aleth-NeRF’s assumption about darkness is grounded in the Concealing Field present in the air, we believe that a more effective modeling of darkness might offer improved solutions for addressing and handling overexposure issues.

\section{Conclusion}

In this paper, we propose Aleth-NeRF model. Inspired by wisdom of ancient Greeks, we introduce a concept: Concealing Field. After incorporating the concealing fields into NeRF, we can effectively address novel view synthesis tasks in scenes with low-light $\&$ over-exposure conditions. Additionally, we proposed several unsupervised loss functions to constrain the generation of the concealing fields, make Aleth-NeRF could achieve SOTA reults on novel view synthesis under low-light $\&$ over-exposure conditions.

One limitation is that Aleth-NeRF should be specifically trained for each scene, which is the same as vanilla NeRF. Besides, Aleth-NeRF may fail in scenes with non-uniform lighting conditions~\cite{wang2022lcdp} or shadow conditions~\cite{shadow}, we believe that this is also a valuable research topic for future exploration.

\section{Acknowledgements}

This work was done during first author internship at Shanghai Artificial Intelligence Laboratory. This work was partially supported by JST Moonshot R$\&$D Grant Number JPMJPS2011, CREST Grant Number JPMJCR2015 and Basic Research Grant (Super AI) of Institute for AI and Beyond of the University of Tokyo. And this work is supported in part by the National Key R$\&$D Program of China (NO. 2022ZD0160100).

Meanwhile, many thanks to the colleagues from the University of Tokyo MIL Laboratory and Shanghai AI Lab for their active discussions in this work. Additionally, special thanks to Tianhan Xu for his valuable suggestions during the dataset collection phase of this work.

\bibliography{aaai24}

\clearpage

\section{Dataset Details}

\begin{figure}[t]
    \centering
    \includegraphics[width=0.95\linewidth]{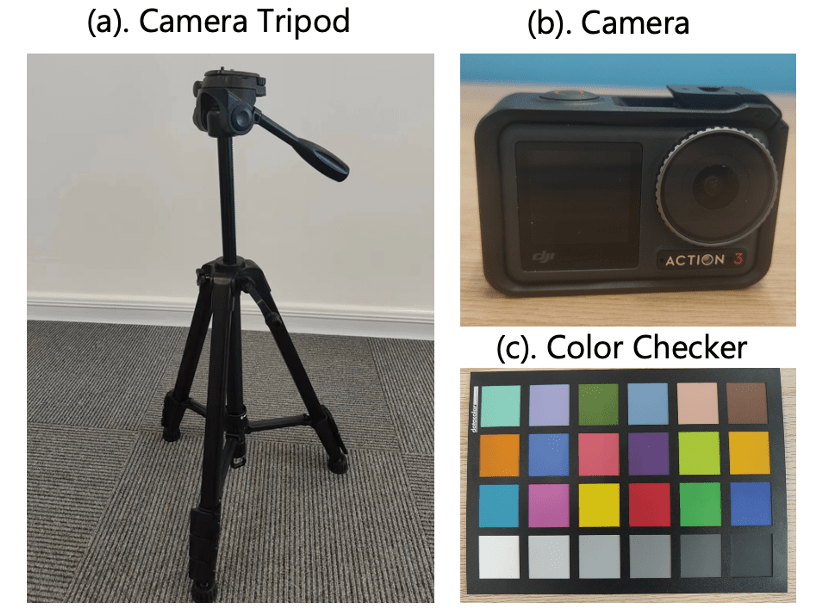}
    \caption{Dataset collection equipment.}
    \label{fig:equipment}
\end{figure}

We captured the scenes using a DJI Osmo Action 3 camera (Fig.\ref{fig:equipment}(b)). To obtain multi-view images, we moved and rotated the camera tripod (Fig.\ref{fig:equipment}(a)). Paired low-light, normal-light, and over-exposure images were created by adjusting the exposure time and ISO value, while maintaining fixed camera configurations, all sRGB images are generated by default camera ISP processor. For precision, each image also featured a DSLR color checker (Fig.~\ref{fig:equipment}(c)) to ensure accurate color representation and facilitate a comprehensive assessment of both Aleth-NeRF-generated images and comparison methods.

\begin{figure*}
    \centering
    \includegraphics[width=1.00\linewidth]{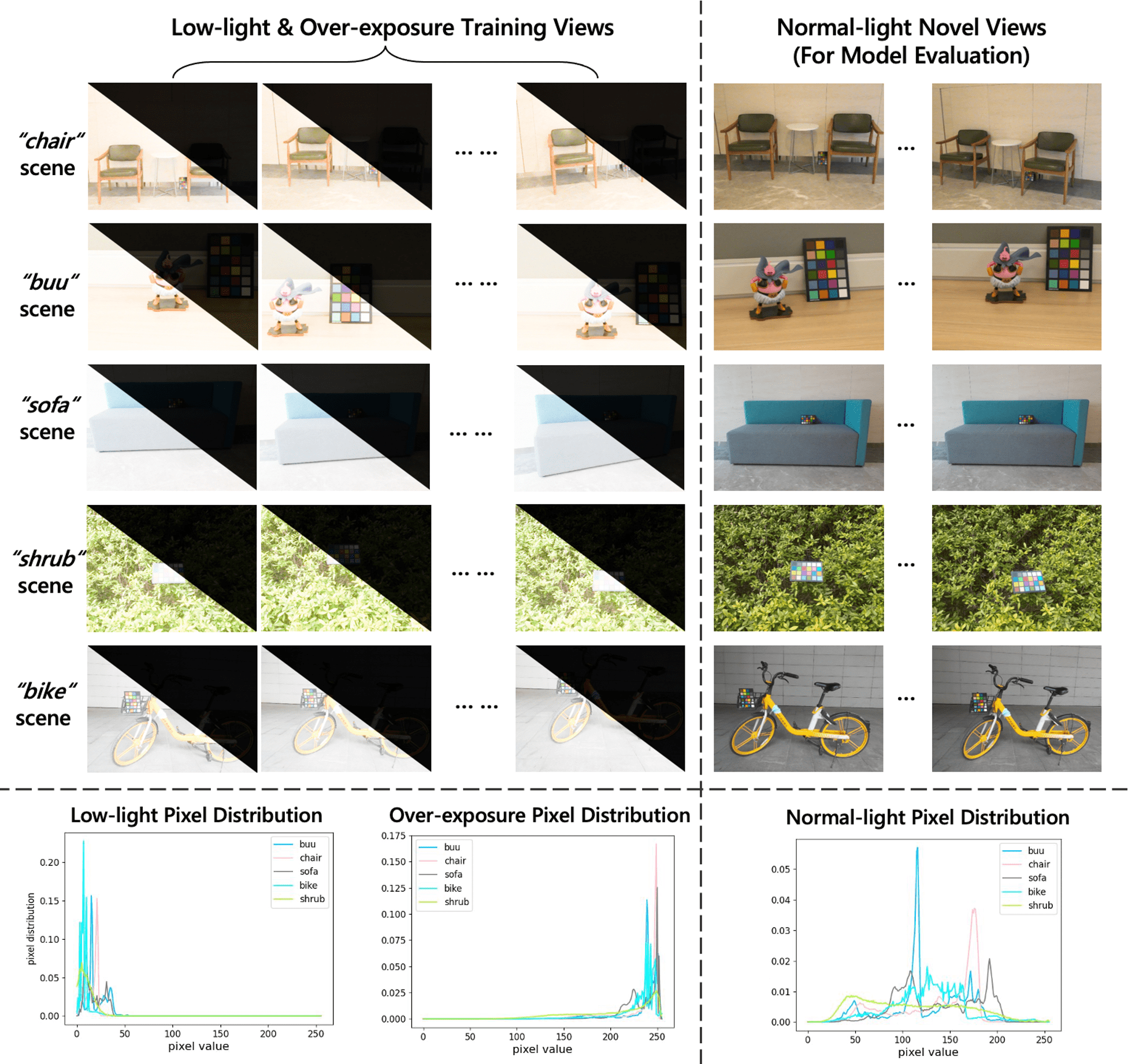}
    \caption{We collected a dataset comprising 5 scenes. Each scene we split the low-light and over-exposure training views used for model training, along with ground-truth normal-light novel views for performance evaluation.}
    \label{fig:dataset}
\end{figure*}

As shown in Fig.~\ref{fig:dataset}, we divide the dataset into training views encompassing low-light $\&$ over-exposure images, and evaluation views representing the ground truth of normal lighting conditions. The number of training and testing views for each scene can be found in Table~\ref{data_split}. The ground truth of the evaluation views we selected is solely for assessing the quality of model generation. In practical usage, Aleth-NeRF is capable of generating seamless and coherent normal light views with given camera position. The Y-channel pixel distribution of low-light $\&$ normal-light $\&$ over-exposure images are shown in the bottom of Fig.~\ref{fig:dataset}, where x-label is the pixel value range from $0 \sim 255$ and y-label is the probability of corresponding pixel value.

\section{Aleth-NeRF on Other Framework}

The Concealing Field assumption of Aleth-NeRF is a broad concept, allowing Aleth-NeRF to be applied to other NeRF~\cite{nerf} follow-up variants like Instant-NGP~\cite{mueller2022instant} to expedite training. To explore this, we conducted an experiment by integrating Aleth-NeRF into the Instant-NGP framework. 

An comparison of low-light condition ``\textbf{\textit{buu}}" scene is shown in Table.~\ref{instantngp}. Within the Instant-NGP framework, the training time for Aleth-NeRF is significantly reduced, although the resulting performance might be slightly inferior compared to the architecture solely based on NeRF. 

\begin{table}[t]
\renewcommand\arraystretch{1.33}
\caption{Aleth-NeRF on different frameworks, low-light condition ``\textbf{\textit{buu}}" scene for example.}
\vspace{-3mm}
\label{instantngp}
\centering
\begin{adjustbox}{max width = 0.90\linewidth}
\begin{tabular}{l|c|c}
\hline
\hline
              & PSNR$\uparrow$/SSIM$\uparrow$/LPIPS$\downarrow$      & Training Time       \\ \hline
+ NeRF        & 19.14/ 0.839 / 0.306 & 2.2hours / 4 GPUs \\ \hline
+ Instant-NGP & 18.09/ 0.806/ 0.344  & 4.8min/ 1GPU      \\ \hline
\hline
\end{tabular}
\end{adjustbox}
\end{table}

\section{More Visualization Results}

\begin{figure*}
    \centering
    \includegraphics[width=0.95\linewidth]{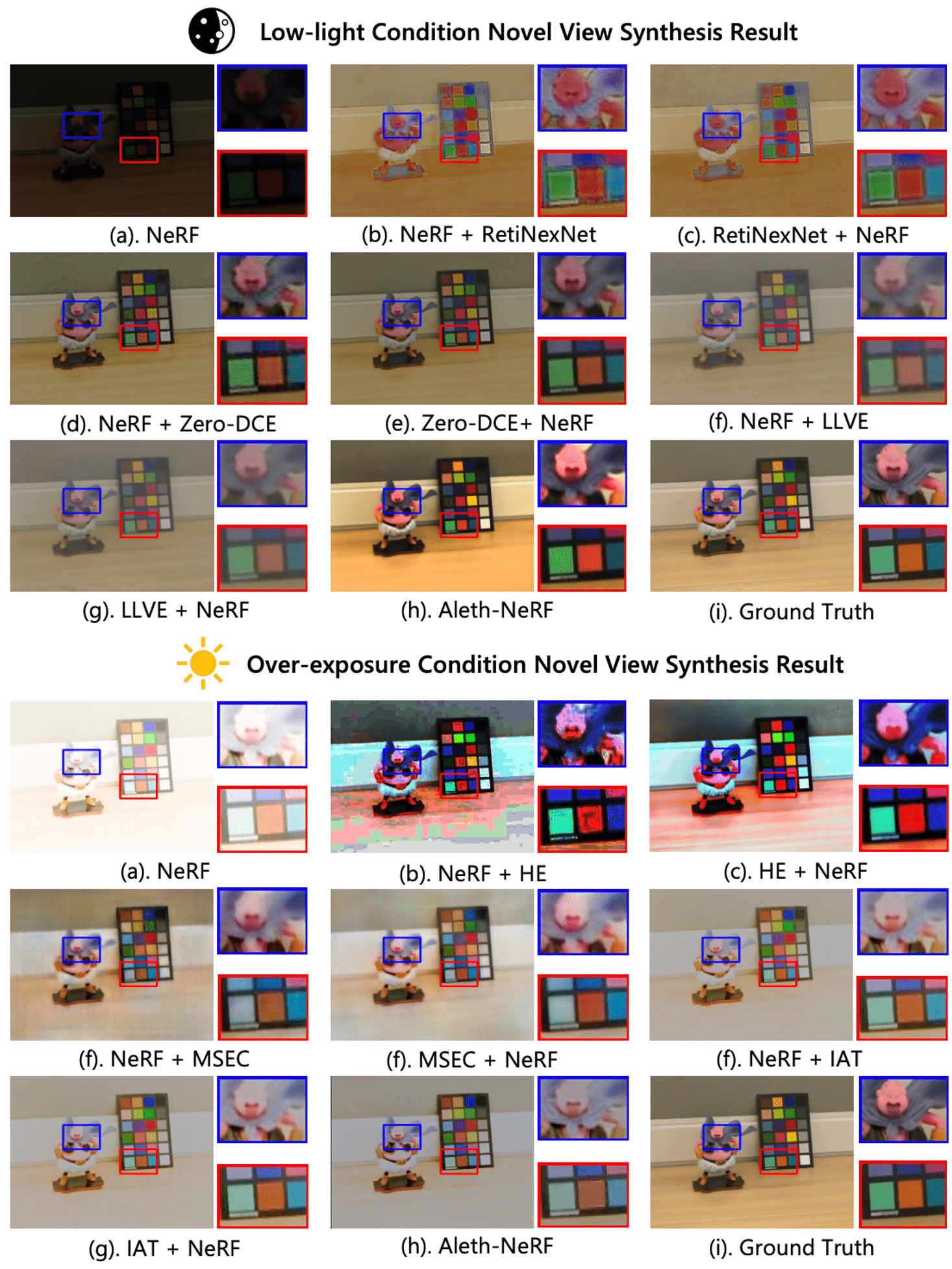}
    \caption{Comparison of novel view generation under low-light $\&$ over-exposure ``\textbf{\textit{buu}}" scene.}
    \label{fig:buu}
\end{figure*}

\begin{figure*}
    \centering
    \includegraphics[width=0.95\linewidth]{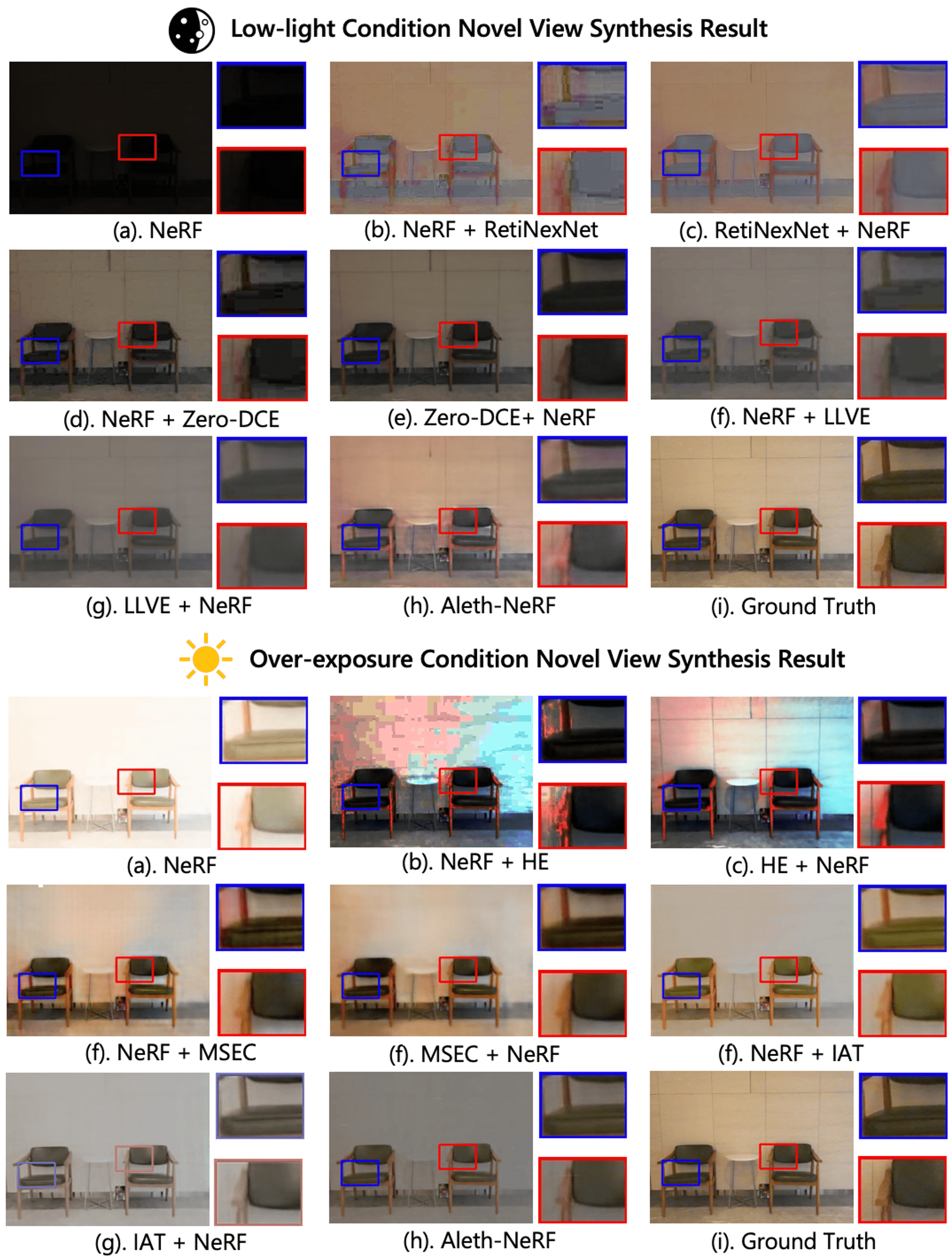}
    \caption{Comparison of novel view generation under low-light $\&$ over-exposure ``\textbf{\textit{chair}}" scene.}
    \label{fig:chair}
\end{figure*}

We show more novel view generation examples of ``\textbf{\textit{buu}}" scene in Fig.~\ref{fig:buu}  and ``\textbf{\textit{chair}}" scene in Fig.~\ref{fig:chair}, these examples encompass rendering results under both low-light and over-exposure conditions.  In panel (a), we show the results of vanilla NeRF rendering.  Panels (b, d, f) correspond to the ``NeRF + *" outlined in Table 1 of the main text, while "*" denotes enhancement methods [(b): RetiNexNet~\cite{RetiNexNet}, (d): Zero-DCE~\cite{Zero-DCE}, (f): LLVE~\cite{LLVE_2021_CVPR})] in low-light scenes and exposure correction methods [(b): histogram equlization~\cite{DIP_citation}, (d): MSEC~\cite{Afifi_2021_CVPR} and (f): IAT~\cite{BMVC22_IAT}] in over-exposure scenes. On the other hand, panels (c, e, g) illustrate the `` + NeRF" strategies as mentioned in Table 1 of the main text, these methods involve enhancing pre-processed training views using enhancement techniques and then training NeRF using the improved views. (h) is the rendering results of our proposed Aleth-NeRF. (g) is the ground truth nomal-light novel view.

We could find that ``NeRF + *" series methods perform worse in image quality when compare with ``* + NeRF” series methods, especially in the low-light condition comparision, this discrepancy implies that NeRF training is susceptible to disruptions caused by low-light and over-exposure conditions, which make the enhancement methods ineffective on the generated views.  ``* + NeRF” series methods denotes to pre-process multi-views with 2D enhancement models, sometimes fail to ensure 3D consistency, leading to blurriness as depicted in Fig.~\ref{fig:buu} and Fig.~\ref{fig:chair}.

\section{Future Research Directions}

We see significant potential in exploring novel view synthesis in challenging lighting conditions. This task involves synthesizing new views and enhancing our understanding of the physical world. Here are some feasible directions:

\begin{itemize}

\item Artificially collecting camera coordinates is labour-intensive, algorithms like COLMAP~\cite{COLMAP1,COLMAP2} often fail in the abnormal lighting conditions. Therefore, the exploration and development of coordinate-free NeRFs are essential.

\item Similar to the original version of NeRF, Aleth-NeRF requires separate training for each scene. Therefore, a key challenge lies in designing a scene-generalizable NeRF that performs well in adverse lighting conditions.

\item Aleth-NeRF can achieve unsupervised lightness correction, while works like NeRF-W~\cite{nerf_wild} can balance NeRF for different lighting conditions. Therefore, exploring how to design more effective physical modeling to simultaneously accomplish both tasks is a valuable direction.

\end{itemize}

\end{document}